\documentclass[10pt,twocolumn,letterpaper]{article}

\usepackage{cvpr}              
\usepackage{microtype}
\usepackage{graphicx}
\usepackage{subcaption}
\usepackage{booktabs} 
\usepackage{placeins}
\usepackage{multirow} 
\usepackage{float}
\restylefloat{table}
\usepackage{hyperref}
\usepackage{siunitx}

\usepackage{graphicx}
\usepackage{amsmath}
\usepackage{amssymb}
\usepackage{booktabs}

\usepackage{mathtools}
\usepackage{algorithm}
\usepackage{algorithmic}
\usepackage{bm}
\usepackage[numbers]{natbib}

\usepackage[capitalize,noabbrev]{cleveref}

\usepackage[textsize=tiny]{todonotes}

\usepackage[capitalize,noabbrev]{cleveref}
\usepackage{hyperref}

\usepackage[capitalize]{cleveref}
\crefname{section}{Sec.}{Secs.}
\Crefname{section}{Section}{Sections}
\Crefname{table}{Table}{Tables}
\crefname{table}{Tab.}{Tabs.}

\def\cvprPaperID{17} 
\def\confName{CVPR}
\def\confYear{2024}

\def\eqref#1{equation~\ref{#1}}

\def\1{\bm{1}}

\def\vtheta{{\bm{\theta}}}

\DeclareMathAlphabet{\mathsfit}{\encodingdefault}{\sfdefault}{m}{sl}
\SetMathAlphabet{\mathsfit}{bold}{\encodingdefault}{\sfdefault}{bx}{n}

\begin{document}

\title{Not Only the Last-Layer Features for Spurious Correlations: All Layer Deep Feature Reweighting}

\author{\parbox{\textwidth}{\centering
    Humza Wajid Hameed$^{1,2}\;$\thanks{ We acknowledge support from FRQNT Grant 2023-NOVA-3979 as well as compute resources provided from Calcul Quebec. Correspondence to: \texttt{humzaw28@gmail.com,
    geraldin.nanfack@concordia.ca,
    eugene.belilovsky@concordia.ca}} \hspace{-10pt}
    \qquad Geraldin Nanfack $^{1,2}$
    \qquad Eugene Belilovsky$^{1,2}$} \vspace{5pt}\\
$^1$ Concordia University, Montreal, QC, Canada; $^2$ Mila -- Quebec AI Institute; 
}
\maketitle

\begin{abstract}
   Spurious correlations are a major source of errors for machine learning models, in particular when aiming for group-level fairness. It has been recently shown that a powerful approach to combat spurious correlations is to re-train the last layer on a balanced validation dataset, isolating robust features for the predictor. However, key attributes can sometimes be discarded by neural networks towards the last layer. In this work, we thus consider retraining a classifier on a set of features derived from all layers. We utilize a recently proposed feature selection strategy to select unbiased features from all the layers. We observe this approach gives significant improvements in worst-group accuracy on several standard benchmarks. \vspace{-15pt}
\end{abstract}

\begin{figure*}
    \centering
    \includegraphics[width=1\linewidth]{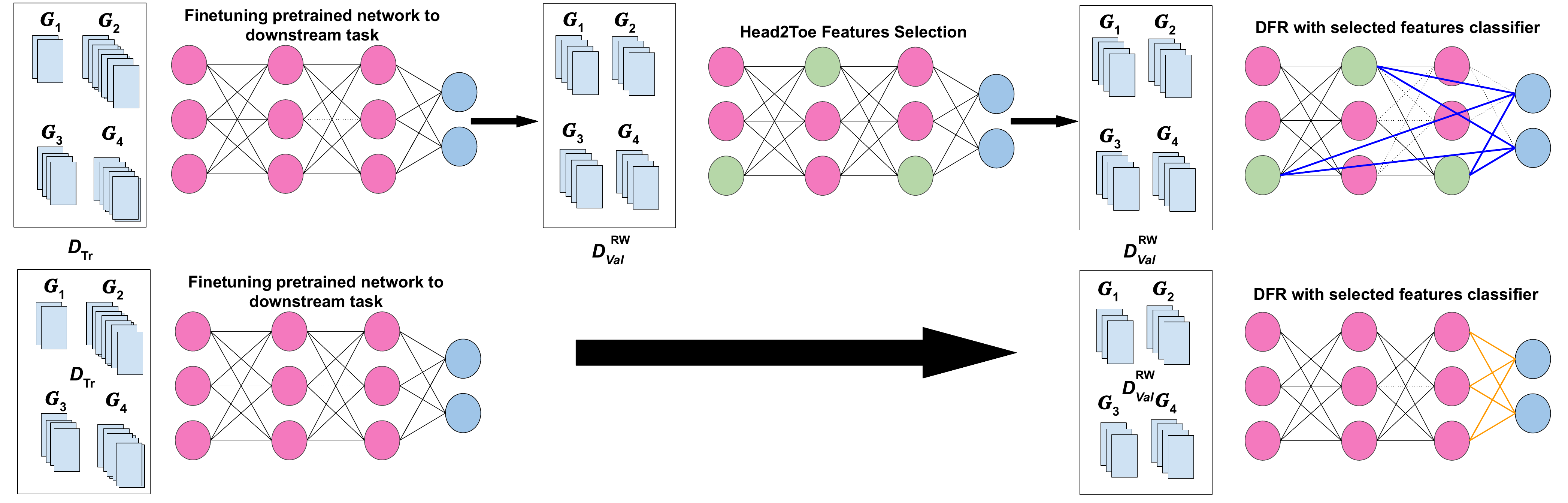}
    \caption{ \textbf{Top Row - }\textbf{H2T-DFR} Illustration of the different training phases for H2T-DFR. A pre-trained network is tuned on a target task using unbalanced data followed by Head2Toe feature selection with balanced training data. Balanced data consists of equal counts of group $G_i$, where each group is a unique combination of target and spurious features. Lastly, a classifier comprised of selected features is trained on an unseen validation dataset. \textbf{Bottom Row} - \textbf{DFR} Illustrates the DFR baseline method in comparison, which excludes any feature selection\vspace{-5pt}}
    \label{fig:spuriousH2T_summary}
\end{figure*}

\section{Introduction}
\label{sec:intro}

Thanks to their performance capability, deep learning is increasingly applied across diverse domains including healthcare. However, when trained with empirical risk minimization (ERM), deep learning models may fail to learn stable features, which are those that hold across data distributions collected at different times and places~\cite{shah2020pitfallssimplicitybiasneural}. For instance, it has been observed the tendency of convolutional neural networks (convnets) to often prioritize image texture over more informative features such as shapes, which may be better predictors~\cite{hermann2020originsprevalencetexturebias,geirhos2020shortcut}.  This tendency arises from models' ability to exploit any shortcuts or spurious correlation present in training data which may be sufficient to correctly predict training data, but may not hold in unseen test data. As a result, this exposure to learning spurious correlation or any shortcut in data makes them vulnerable to a potential drop in predictive performance.

Addressing the challenge of learning in the presence of such spurious correlations has motivated several approaches in the literature. It is widely adopted the notion of \textit{groups} (defined in Sec.~\ref{sec:background_and_related_work}), in which to correctly classify instances from certain groups, it is expected that models should be robust enough to spurious correlation. Assuming the presence of group annotations in training data, a well-known alternative to ERM is group distributionally robust optimization (group DRO)~\cite{sagawadistributionally}, which minimizes the empirical worst-group risk. There exist methods that do not assume group information during training such as just train twice (JTT) \cite{liu2021just}, which divides the training time into two phases. The first phase trains the model with ERM, while the second phase continues by up-weighing the loss of misclassified instances of the first phase. Recently, it has been shown that the last-layer representations of ERM-trained models already exhibit both robust and non-robust (to spurious correlation) features~\cite{kirichenko2022last,izmailov2022feature}. As a remedy, to decrease the impact on non-robust features, deep feature reweighting (DFR) \cite{kirichenko2022last} has proven effective, which only retrains the classifier with a balanced-group validation set. 

DFR can be viewed as an instantiation of transfer learning, where there is a desire to exploit robust and generalizing features from the source domain to build a good predictive model on the target domain~\cite{tan2018survey}. Here, the goal of the target domain is a dataset balanced according to groups. In transfer learning literature, more advanced methods exist beyond retraining the classifier. Indeed during supervised learning a network can learn to discard certain robust features present from earlier layers to make the final prediction in the last layer, thus losing potentially useful features. This paper considers a simple yet efficient transfer learning method, called Head2Toe~\cite{evci2022head2toe}.  Unlike last-layer retraining, Head2Toe leverages all layer features, not just the last one, to find a sparse network with the most transferable features. Therefore, we leverage Head2Toe to complement DFR and aim to get the most transferable features while also decreasing the impact of non-robust (to spurious correlation) features. 
   
Our contributions can be summarized as follows: (i) we show how an efficient transfer learning method (Head2Toe here) can be incorporated in a pipeline of a state-of-the-art method in spurious correlation learning, (ii) we demonstrate that this incorporation can yield better performance on standard evaluation benchmarks.
\vspace{-3pt}
\section{Related Work}
Several approaches have been developed to address the challenge of learning amidst spurious correlations, alongside various methods aimed at efficient transfer learning.

\noindent\textbf{Learning with Spurious Correlation.}
Data augmentation techniques appear to be the standard approach to fight against minority groups~\cite{agarwal2020towards,chen2020counterfactual,shetty2019not}. For example, \cite{plumb2022findingfixingspuriouspatterns} introduces a data augmentation technique by generating counterfactual data, which adds or removes object parts responsible for identified spurious patterns. 
There exist methods that analyze representations throughout the training dynamics to understand how the bias to spurious features arises. For instance,
\cite{dreyer2023hope} propose a method to reduce model reliance on spurious features by penalizing the gradient in directions of spurious features. Several other works have been done to analyze stochastic gradient descent (SGD) directions, leading to modified versions of SGD or loss functions~\cite{pezeshki2021gradient,rahaman2019spectral,nagarajanunderstanding}. Although most of these methods address the spurious correlation problem, it has been recently shown the ability of ERM to competitively learn robust-to-spurious-correlation features~\cite{kirichenko2022last,zong2022medfair}. 
Despite the use of data with labeled spurious features in DFR, it differs from the above works by directly using balanced data where each grouping of class label and spurious feature are equally represented, resulting in a decrease in bias toward spurious or unstable features.

\noindent\textbf{Efficient Transfer Learning.}
Transfer learning still has its challenges addressed across various works and there are differing techniques established to improve downstream task performance.
\cite{yazdanpanah2022revisiting} presents a method that consists of training affine parameters in batch normalization layers. It explains how training these shift and scale parameters used in the normalization step can have a noticeable impact, especially in scarce data settings.
Downstream tasks where there is minimal divergence in distribution from the pre-trained model seldom have difficulty learning on a new dataset when employing finetuning. This assumes that the pre-trained model is otherwise sufficient for the downstream task and improvements can be attained through leveraging batch normalizations alone. \cite{kumar2021fine} explains that these in-distribution tasks struggle most when exposed to data with significant deviations in distribution from the source task. It explains that with a randomly initialized linear head, finetuning results in larger fluctuations in parameter weights during the earlier training steps as the whole network needs to adjust to what it considers an unusual dataset relative to the source task. First keeping the pretrained backbone frozen and training the linear head before unfreezing and finetuning allows the model to better adapt and transfer to out-of-distribution downstream tasks. Here, instead of drawing focus to just one layer, the assumption is that a better initialization of the classifier will lead to improved finetuning on the downstream task.
\cite{qiu2023simple} deals with the out-of-distribution transfer learning problem through the use of a weighted loss function. It proposes Automatic Feature Reweighting (AFR), a method that reweighs the loss among groups with fewer examples to push the model to better adjust and adapt for these minority group examples. This method has the additional benefit of not relying on spurious features. This approach leans similar to the DFR method by rebalancing groups to avoid unstable features from dominating the learning process although it does it through a loss function instead of group counts.
With these different approaches to efficient transfer learning, the motivation for our method is derived from the use of Head2Toe’s unique feature selection process. Unlike the above-mentioned methods, Head2Toe focuses on searching for information found throughout the network instead of relying solely on the penultimate layer. Head2Toe extracts useful intermediate layer features \cite{evci2022head2toe} and concatenates them to create a linear layer. This newly initialized linear layer, combined with Deep Feature Reweighting (DFR) \cite{kirichenko2022last}, is trained on an unseen balanced dataset under our approach. With Head2Toe improving feature extraction by incorporating early layer information and DFR reducing unstable feature bias through balanced data training, our method is motivated to address challenges in efficient transfer learning.
   
\section{Background}\label{sec:background_and_related_work}
 We consider a classification problem with a training set denoted by $\mathcal{D}_{\text{Tr}} = \{(\mathbf{x}_i, y_i)\}_{i=1}^N$, where $\mathbf{x}_i \in \mathcal{X}$ is the input and $y_i\in \mathcal{Y}$ is its class label. For each data $\mathbf{x}_i$, there is a spurious attribute value $a_i \in \mathcal{A}$ of $\mathbf{x}_i$, where $a_i$ is non-predictive of $y_i$, and $\mathcal{A}$ denotes the set of all possible spurious attribute values. We denote by a group the pair $g:= (y,a) \in \mathcal{Y}\times \mathcal{A}:= G$. 
Our goal is to learn a parameterized model~\footnote{
To simplify, we can omit subscripting parameters $\vtheta$.}: $f_{\theta}: \mathcal{X}\longrightarrow \mathcal{Y}$ that will maximize the expected accuracy while avoiding learning spurious features. We consider a neural network model \begin{equation}
 f_{\theta}(.) = g\left( h(.)\right),
  \label{eq:neuralNetModel}
\end{equation} where $g$ denotes its classifier layer and $h$ denotes its feature network. 
    Additionally, we denote by $\mathcal{D}_{\text{Val}}$ and $\mathcal{D}_{\text{Te}}$ the validation and test sets, respectively. Assuming that we have the information of groups on a sample $\mathcal{D}$, we denote by $\mathcal{D}^{\text{RW}}$ a balanced subset of $\mathcal{D}$ in which groups are uniformly distributed, i.e., each group has the same number of examples.

\noindent\textbf{Deep Feature Reweighting.}
Deep feature reweighting (DFR) \cite{kirichenko2022last} is a method to reduce bias towards spurious features, which are features in the data that are non-predictive of the task yet may be highly correlated with training targets. DFR training involves two phases: initially, the model is trained using empirical risk minimization (ERM) on the training data $\mathcal{D}_{\text{Tr}}$, without the information of the spurious attribute. Second, the feature network $h$
is frozen, and the classifier $g$ is trained on a balanced validation set $\mathcal{D}_{\text{Val}}^{\text{RW}}$. The core idea in DFR is that ERM learns both robust and non-robust (to spurious correlation) features, and the second phase enables prioritization of stable features through knowledge of group annotations.

\noindent\textbf{Head2Toe.}
Head2Toe \cite{evci2022head2toe} is one of the techniques for efficient transfer learning. It aims to select the most useful features from intermediate layers that better transfer to a target or downstream task. Head2Toe starts by concatenating all feature maps for intermediate layers, then learns a linear head on top of all these feature maps using the group-lasso regularizer \cite{yuan2006model}. This group-lasso regularizer allows the computation of relevant scores based on each feature's $l_2$ norm of weights. Using a threshold $\tau$, one can select the most useful features that better transfer to the downstream task. It has been shown that this method for transfer learning usually performs better than only retraining the linear head. In the following sections, we denote by $h^{\text{H2T}}$ the feature network found by the Head2Toe method.

\section{Method}\label{sec:method}
This section describes our method H2T-DFR, which is summarized in Figure~\ref{fig:spuriousH2T_summary} and Algorithm 1. It consists of three phases detailed below: (1) unbalanced training (or finetuning) on all the training data (2) balanced data feature selection and finally using the features for (3) balanced data linear classifier training. 
\begin{figure}
    \centering
    \includegraphics[width=1\linewidth]{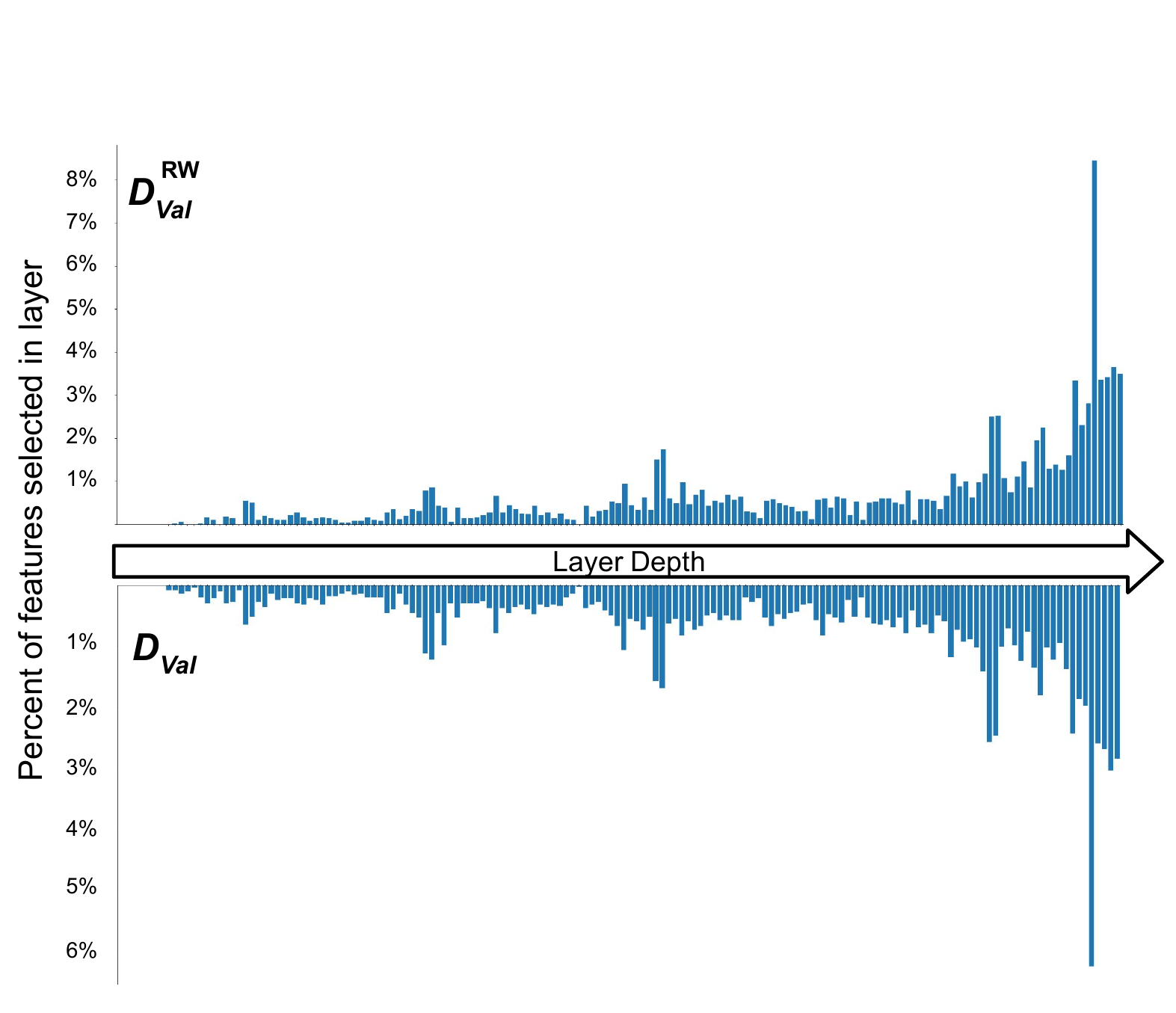}
    \caption{\textbf{HAM10000 and \textbf{H2T-DFR}}: Layerwise proportion of features extracted among the overall top 5\% features selected using balanced \textbf{(}\textbf{top)} and unbalanced data \textbf{(bottom}\textbf{)} for feature selection. Layer depth displayed on the x-axis illustrates a minimal amount of features selected from early layers while most attributes are retained from deeper layers.}
    \label{fig:features_from_layers}
\end{figure}

\begin{table*}[t]
\small
    \centering
    \begin{tabular}{|l|c|c|c|} \cline{3-4}
          \multicolumn{2}{c|}{}&  Worst-group accuracy& Mean group accuracy\\ \hline 
          Dataset&Method&  Mean (\%)&  Mean (\%)\\ \hline 
          &DFR&  $85.99 \pm 0.74$ &  $91.58 \pm 0.15$\\ 
          CelebA&Affine-DFR&  $85.49 \pm 0.70$&  $91.55 \pm 0.14$\\ 
          &H2T-DFR (ours)&  $\bm{88.59 \pm 0.48}$&  $\bm{91.87 \pm 0.17}$\\ \hline 
          &DFR
&  $\bm{92.76 \pm 0.37}$&  $\bm{94.54 \pm 0.21}$\\ 
          Waterbirds&Affine-DFR
&  $89.02 \pm 0.37$&  $94.13 \pm 0.08$\\  
          &H2T-DFR (ours)&  $90.85 \pm 0.45$& $ 93.51 \pm 0.06$\\ \hline 
          &DFR& $ 67.31 \pm 2.61$ & $ 78.09 \pm 0.91$\\
          HAM10000&Affine-DFR& $ 53.63 \pm 2.84 $ &  $76.72 \pm 0.68$\\ 
          &H2T-DFR (ours)&  $\bm{69.69 \pm 1.84}$& $ \bm{78.23 \pm 0.46}$\\ \hline
    \end{tabular}
    \caption{Mean and worst-group performance over 5 seeds. Mean group accuracy averages over all group accuracies.\vspace{-8pt}}
    \label{tab:accuracies}
\end{table*}

\noindent\textbf{Unbalanced Finetuning.}
As in the DFR approach our method starts by simple fine-tuning on the entire dataset using ERM on $\mathcal{D}_{\text{Tr}}$, denoting this trained backbone as $h_{\text{pretrained}}$. 

\noindent\textbf{Balanced Feature Selection and Classifier Training.} 
 We follow an approach similar to Head2Toe: we concatenate all the features from all layers and perform a group lasso feature selection. Using $h_{\text{pretrained}}$, we initialize a Head2Toe model which will have the linear layer $g_{\text{H2T}}$. It should be noted that $g_{\text{H2T}}$ in this stage is a linear classifier layer with inputs from all intermediate features in the network as described in the Head2Toe background section.  
 $g_{\text{H2T}}$ is then trained on $\mathcal{D}_{\text{Val}}^{\text{RW}}$ using a group-lasso regularization loss. The regularization allows the model to adjust weights in terms of importance, and the resulting weights can be used to calculate \textit{relevance scores} $s_i$ \cite{evci2022head2toe}, where $s_i$ are the $l_2$ norms of the final weights of each feature of the $h_{\text{pretrained}}$ network. The final features selected correspond to the ones with the top $\tau$ percent scores, and the corresponding sparse network is denoted by $h_{\text{H2T}}$.  
    
    With features selected, our final model is initialized using the sparse network $h_{\text{H2T}}$, which contains only selected features instead of all features from the pre-trained network. We denoted by $f_{\text{H2T}} = g_{\text{H2T}} \circ h_{\text{H2T}}$ the full head2Toe model alongside its classifier.
    It should be noted that the classifier here, $g_{\text{H2T}}$, is a new linear layer with inputs $h_{\text{H2T}}$ features selected in Part 1. The backbone is frozen and $g_{\text{H2T}}$ trained on $\mathcal{D}_{\text{Val}}^{\text{RW}}$, where each batch has equal group counts and removes the ability for the model to build a bias towards minority groups. The resulting model is what we denote \textbf{H2T-DFR}. The difference here being that although $\mathcal{D}_{\text{Val}}^{\text{RW}}$ has been used for feature selection previously and is composed of equal group counts in each batch, more importantly, it has never been introduced to the model for parameter updates. During our experiments, we observed that a balanced dataset alone does not improve results. It is also required that the dataset has never been previously used to train the model as in DFR. 

\section{Experiments}

We now discuss our experimental results. We focus our comparison to DFR which has shown to be a powerful technique that can outperform existing methods such as groupDRO~\cite{izmailov2022feature}. We propose another baseline to compare to DFR and H2T-DFR, specifically \textbf{Affine-DFR}. This approach mimics DFR and does not have an explicit feature selection phase. Similar to DFR, \textbf{Affine-DFR} uses a balanced training phase but adapts only the affine parameters of batch-norm layers.
    
Experiments for DFR, Affine-DFR and H2T-DFR were run over 5 seeds and table \ref{tab:accuracies} shows their respective performance on CelebA~\cite{liu2015deep}, Waterbirds~\cite{sagawadistributionally}, and HAM10000~\cite{zongmedfair,hermann2020originsprevalencetexturebias}. The pre-trained model employed is ResNet-50~\cite{he2016deep}. Hyperparameters were selected to reproduce result from the respective papers employed. 
For spuriousH2T, there is further hyperparameter tuning done starting from the hyperparameters in table \ref{tab:SpuriousH2THyperparameters} as there is added complexities such as feature selection fraction and layer target size to consider along with the different learning rates across the 3 training phases. Hyperparameters were originally set according to papers referenced for baseline results and further tuned through a hyperparameter search similar to the DFR paper \cite{kirichenko2022last}.

\noindent\textbf{Results.}
Our results shown in Tab.~\ref{tab:accuracies} are with the use of a balanced dataset for feature selection. We observe improvement in the worst group accuracy for HAM10000 and CelebA, whereas
Waterbirds does not show improvement from H2T features, the result being very close to that of the basic DFR. We note however that this dataset is known to be simpler than the others considered. Indeed, for Waterbirds, it was already observed that ImageNet pre-trained models already contain robust features to easily get $\approx 88\%$ of worst-group accuracy by just retraining the classifier \cite{izmailov2022feature}. CelebA's worst-group accuracy increased by 2.60\% from our method relative to the baseline (DFR). The medical dataset HAM10000 shows a 2.38\% increase in worst-group accuracy. With all 3 datasets, the mean across group accuracies did not show a noticeable difference between the 3 methods. 

To gain insight into the feature selection we illustrate the features selected depending on depth for HAM1000 in figure \ref{fig:features_from_layers} (but note similar trends are observed in the other datasets). This illustrates that despite not relying solely on the penultimate layer for feature selection, the most crucial information is found towards the end layers of the network. We also compare to a feature selection done on unbalanced data and observe that this tends to select more features from lower layers, which we attribute to the presence of strong spurious features in lower layers that may be selected for the unbalanced problem setting. 

\vspace{-3pt}
\section{Conclusion}\label{sec:conclusion}
This paper studies the problem of learning in the presence of spurious correlation. We propose H2T-DFR, a three-stage method that leverages Head2Toe (an efficient transfer learning method)~\cite{evci2022head2toe}, and incorporate it in the pipeline of DFR~\cite{kirichenko2022last}, a state-of-the-art method to fight against spurious correlation. H2T-DFR selects the most transferable features from all layers before applying DFR. Experiments on standard evaluation benchmarks demonstrate that H2T-DFR improves DFR, showing that efficient transfer learning methods can boost the worst-group predictive performance of robust-to-spurious correlation methods. 




\newpage
{\small
\bibliographystyle{ieee_fullname}
\bibliography{PaperForReview}
}
\newpage

\appendix
\onecolumn
\section{Appendix}
Here we provide the Algorithm detailing the H2T-DFR method in Algorithm 1. Finally, Tables 2-4 describe the hyperparameters used in our experiments for the different baselines.

\begin{algorithm*}[t]
   \caption{Spurious Head2Toe (\textbf{H2T-DFR})}
   \label{alg:spuriousH2T}
    \begin{algorithmic}

   \STATE {\bfseries Input:} training, validation, testing sets: $\mathcal{D}_{\text{Tr}}$, $\mathcal{D}_{\text{Val}}$, $\mathcal{D}_{\text{Te}}$; percentage of selected features $\tau$;  
   number of epochs $T_{1}$ and $T_2$ and $T_\text{DFR}$. 
   
   \textit{\textbf{Finetuning for a downstream task}}
   \begin{enumerate}
    \STATE Initialize model with pre-trained model $f= g \circ h$
    \STATE Finetune $g$ and $h$ on $\mathcal{D}_{\text{Tr}}$  for $T_{1}$ epochs
   \end{enumerate}
  \text{\textit{\textbf{Feature Selection}}}
   \begin{enumerate}
       \STATE Initialize the Head2Toe model $f_{\text{H2T}}$ with classifier consisting of all layers 
       \STATE Get the set of most important features $\mathcal{H_{FS}}$  using group lasso \cite{evci2022head2toe} on $\mathcal{D}_{\text{Val}}^{\text{RW}}$ with $T_2$ epochs (see Section~\ref{sec:method})
   \end{enumerate}

    \text{\textit{\textbf{Deep Feature Reweighting: Balanced data retraining}}} 
     \begin{enumerate}
         \STATE Initialize the $f_\text{H2T}$ through $h_{\text{H2T}}$ with only selected features $\mathcal{H_{FS}}$ 
    \STATE Freeze $h_\text{H2T}$ and retrain the classifier $g_\text{H2T}$ on $\mathcal{D}_{\text{Val}}^{\text{RW}}$ for $T_\text{DFR}$ epochs
    \end{enumerate}
    \end{algorithmic}
\end{algorithm*}

\begin{table}[t]
    \centering
    \begin{tabular}{|l|c|c|l|} \hline 
          &Waterbirds& CelebA&HAM10000\\ \hline 
          Optimizer&SGD&  SGD&SGD\\ \hline 
          Learning Rate&0.003&  0.0005&0.0003\\ \hline 
          Weight Decay&0.0004&  0.0001&0.0001\\ \hline 
          Momentum&0.9&  0.9&0.9\\ \hline 
          DFR Learning Rate&0.0001&  0.0001&0.0005\\ \hline 
          DFR Weight Decay&0.0001&  0.0001&0.0004\\ \hline 
          DFR Momentum&0.9&  0.4&0.9\\ \hline 
 DFR Optimizer& SGD& SGD&SGD\\ \hline 
 Epochs& 20& 6&100\\ \hline 
 DFR Epochs& 100& 50&500\\ \hline 
 Batch Size& 32& 128&128\\ \hline
    \end{tabular}
    \caption{DFR - Hyperparameter setting for the 3 datasets presented.}
    \label{tab:SpuriousLinHyperparameters}
\end{table}

\begin{table}[b]
    \centering
    \begin{tabular}{|l|c|c|l|} \hline 
          &Waterbirds& CelebA&HAM10000\\ \hline 
          Optimizer&SGD&  SGD&SGD\\ \hline 
          Learning Rate&0.003&  0.0005&0.0003\\ \hline 
          Weight Decay&0.0004&  0.0001&0.0001\\ \hline 
          Momentum&0.9&  0.9&0.9\\ \hline 
          DFR Learning Rate&0.0001&  0.0001&0.0005\\ \hline 
          DFR Weight Decay&0.0001&  0.0001&0.0004\\ \hline 
          DFR Momentum&0.9&  0.4&0.9\\ \hline 
 DFR Optimizer& SGD& SGD&SGD\\ \hline 
 Epochs& 20& 6&100\\ \hline 
 DFR Epochs& 100& 50&500\\ \hline 
 Batch Size& 32& 128&128\\ \hline
    \end{tabular}
    \caption{Affine-DFR - Hyperparameter setting for the 3 datasets presented}
    \label{tab:Affine-DFRHyperparameters}
\end{table}

\begin{table}[t]
    \centering
    \begin{tabular}{|l|c|c|l|} \hline 
          &Waterbirds& CelebA&HAM10000\\ \hline 
          Optimizer&SGD&  SGD&SGD\\ \hline 
          Learning Rate&0.0005&  0.0005&0.0003\\ \hline 
          Weight Decay&0.0004&  0.0001&0.0001\\ \hline 
          Momentum&0.9&  0.9&0.9\\ \hline 
          DFR Learning Rate&0.0005&  0.0005&0.0005\\ \hline 
          DFR Weight Decay&0.0003&  0.0003&0.0004\\ \hline 
          DFR Momentum&0.9&  0.9&0.9\\ \hline 
 DFR Optimizer& SGD& SGD&SGD\\ \hline 
 Regularization Coefficient& 0.00001& 0.00001&0.0001\\ \hline 
 Epochs& 70& 20&100\\ \hline 
 DFR Epochs& 500& 300&500\\ \hline 
 Batch Size& 32& 128&128\\
    \end{tabular}
    \caption{H2T-DFR - Hyperparameter setting for the 3 datasets presented}
    \label{tab:SpuriousH2THyperparameters}
\end{table}

\end{document}